\documentclass{article}

\usepackage{arxiv}
\usepackage{booktabs}
\usepackage[utf8]{inputenc} 
\usepackage[T1]{fontenc}    
\usepackage{hyperref}       
\usepackage{url}            
\usepackage{booktabs}       
\usepackage{amsfonts}       
\usepackage{nicefrac}       
\usepackage{microtype}      
\usepackage{lipsum}
\usepackage{fancyhdr}       
\usepackage{graphicx}       
\graphicspath{{media/}}     
\usepackage{longtable,booktabs,array}
\usepackage{orcidlink}
\usepackage{authblk}
\usepackage{float}
\title{AI training resources for GLAM: a snapshot}
\author[4]{Andrew Darby}
\author[2]{Catherine Nicole Coleman \orcidlink{0000-0002-3360-1975}}
\author[2]{Claudia Engel}
\author[1]{Daniel van Strien \orcidlink{0000-0003-1684-6556}}
\author[3]{Mike Trizna \orcidlink{0000-0002-0537-8382}}
\author[2]{Zachary W. Painter \orcidlink{0000-0002-4588-7843}}
\affil[1]{British Library}
\affil[2]{Stanford Libraries}
\affil[3]{Smithsonian}
\affil[4]{University of Miami Libraries}

\begin{document}
\maketitle

\begin{abstract}
We take a snapshot of current resources available for
teaching and learning AI with a focus on the Galleries, Libraries,
Archives and Museums (GLAM) community. The review was carried out during 2021 and 2022. The review provides an overview of material we identified as being relevant, offers a description of this material and makes recommendations for future work in this area. \footnote{This review was collated as part of the ai4lam Teaching and Learning Working Group. All  authors made an equal contribution to the paper.}
\end{abstract}


\tableofcontents

\section{Introduction}

In this review, we take a snapshot of current resources available for
teaching and learning AI with a focus on the Galleries, Libraries,
Archives and Museums (GLAM) community. In particular, we seek to assess
the content and topics covered by the material, the methods of delivery
used, the target audience and the current maintenance state of the
materials. Most importantly we ask: How useful and relevant is the
material for the GLAM community? Our objective is to identify to what
extent existing materials, efforts, and approaches are useful and
relevant for the GLAM community and to determine gaps where additional
efforts are needed.

We argue that AI skill building in the GLAM sector is crucial for the
following reasons:

\subsection{GLAM staff need to be able to apply machine learning in their
organizations appropriately and critically.}

Machine learning and artificial intelligence (AI) are rapidly developing
both as academic fields of research and as tools used in software and
services across the public and private sectors. There is an increasing
interest in using AI and machine learning in the GLAM sector\footnote{As
  identified in various recent reports, for example:
 
 \href{https://files.eric.ed.gov/fulltext/ED603715.pdf}{Responsible Operations:
Data Science, Machine Learning,
and AI in Libraries},  \href{https://labs.loc.gov/static/labs/work/reports/Cordell-LOC-ML-report.pdf}{Machine Learning + Libraries}, \href{https://pro.europeana.eu/files/Europeana_Professional/Europeana_Network/Europeana_Network_Task_Forces/Final_reports/AI\%20in\%20relation\%20to\%20GLAMs\%20Task\%20Force\%20Report.pdf}{AI IN RELATION TO GLAMS TASK FORCE}},
but there are challenges that may be unique to GLAMs. An absence of
domain expertise and influence has been identified as one of the reasons
that some of the academic advances in machine learning have not
`translated' into practical applications. For example, a recent review
paper assessed machine learning work related to the detection and
prognosis of COVID-19. Of the 64 studies included in the review after
screening, they found ``that none of the models identified are of
potential clinical use due to methodological flaws and/or underlying
biases.''\footnote{\url{https://www.nature.com/articles/s42256-021-00307-0}}
Although the stakes may not usually be as high\footnote{Whilst GLAM data  is unlikely to be regularly used as the basis of clinical decision
  making there are other areas with high stakes where GLAM data is
  involved for example in historic climate research
  \url{https://www.zooniverse.org/projects/drewdeepsouth/southern-weather-discovery}},
in the GLAM setting the potential benefits of AI/machine learning for
the sector may not be fully realized if GLAM staff do not have the
expertise to shape the application of machine learning.

\subsection{GLAM staff need to be equipped to meet the challenges of working
with their collections.}

GLAM collections pose a range of challenges when used with modern
machine learning methods and tools. GLAM organizations have increasingly
looked at AI to improve the often overwhelming task of accessioning and
processing archival material and making it digitally available. There
are also specific `technical' challenges, for example effectively
applying OCR to historic fonts with poorly preserved materials. Other
challenges come from working with collections that are closely linked to
what gets archived and what eludes collection and preservation. Whilst
collections can be comprehensive, the materials are often not
representative and can significantly shape downstream models based on
these collections. As an example, British collections related to the
history of India are likely to be skewed towards English and Hindi
language materials in comparison to other languages spoken in India.
Administrators are looking for strategies to integrate AI into existing
GLAM technologies and workflows, including catalogues and search
interfaces. Similarly, administrators often face the challenge of
institutional buy-in and access to staff and funding, which requires
their ability to make a good case for AI at their institution.

\subsection{GLAM staff are uniquely positioned to shape the ethical use of
machine learning and AI beyond their organizations.}

GLAM curators and archivists are experts in the selection, description,
preservation, and access to archival materials, including those that are
increasingly digital. A keen awareness of context informs GLAM practices
of metadata development (classification) as well as collection
development. The attention to provenance, in particular, helps place
sources historically, socially, and politically and reveal patterns and
absences, and this attention is sorely needed when it comes to the
preparation of training data for machine learning.

The AI community has also struggled, and too often failed, to address
concerns of privacy, protection of intellectual property, transparency,
and democratization, all of which are core values that GLAM
organizations have been developed to address. \hspace{0pt}In the
increasingly ethically challenged world of machine learning, GLAM staff
are not only ideally positioned but have an obligation to bring their
considerable experience and expertise to bear as stewards of training
data -\/- its creation, documentation, preservation, and reuse.

Part of the inspiration to perform this review goes back to the
recommendations of two prominent reports on AI in GLAMs related to
training. Cordell in the Machine Learning + Libraries
\href{https://labs.loc.gov/static/labs/work/reports/Cordell-LOC-ML-report.pdf?loclr=blogsig}{report}
(pp.60-63) recommends to:

\begin{itemize}
\item
  \begin{quote}
  Develop Modules for ML Training in MLIS Programs (section 5.5.2.)
  \end{quote}
\item
  \begin{quote}
  Cultivate Opportunities for Professional Development in ML (section
  5.5.3.)
  \end{quote}
\item
  \begin{quote}
  Develop Guidelines for Vendor Solutions (section 5.5.5.)
  \end{quote}
\end{itemize}

Padilla in
\href{https://www.oclc.org/content/dam/research/publications/2019/oclcresearch-responsible-operations-data-science-machine-learning-ai.pdf}{{Responsible
Operations}} (p.18) suggests to:

\begin{quote}
1. Initiate evidence-based evaluations of existing data science, machine
learning, and AI training opportunities within and outside of the
library community.

2. Pilot and/or support the development of evidence-based data science,
machine learning, and/or AI training options that are grounded in
library use cases.

3. Explore, document, develop, and share sustainability models for
keeping training opportunities free or low cost without sacrificing
quality and fair compensation.
\end{quote}

These two reports also have helped inform the `gaps' we identify below
and our recommendations. Appropriate professional development and
training may partially address the challenges faced by GLAM institutions
and mentioned in these reports.

\section{Method}

This review had two main goals:
\begin{itemize}
    \item to gather a collection of teaching and learning materials for
machine learning and
    \item to carry out an evaluation of these materials.
\end{itemize}

The collection is based on Google searches\footnote{Discovery can be a
  bit challenging. For example, searching for `machine learning +
  libraries' did not produce relevant results.} and GitHub searches as
well as materials the members of this working group were familiar with.
We primarily collected resources with potential relevance for GLAMs, but
we also included some courses with a broader focus. It was beyond the
scope of this review to capture all possible resources. It was also not
feasible to work through the materials in detail and evaluate their
quality. Instead, we aimed to provide a high level snapshot of current
material and to identify strengths and limitations within a broader
ecosystem. At the point of writing our
\href{https://docs.google.com/spreadsheets/d/1URNmp3RnIP_ppYNwiQeN92cHx2nmZcvMh_UL4E-YyG8/edit?usp=sharing}{{spreadsheet}}
contains 28 sources.

We assessed each teaching resource according to: topics covered, type of
resource, study time required, creator/author, date published, updates,
stated audiences, training data used, and whether the course had an
explicit GLAM focus, either using GLAM collections or GLAM being
mentioned as a target audience.

\section{Findings} 

\subsection{Format}

The following table breaks down the resources by type. Perhaps not
surprisingly, more than half of the materials are based on Python and
Notebooks. A few require running commands in a shell. About 40\% of the
materials consist of Web based texts. 25\% include video clips and
lectures and very few provide slides. We also identified 4 full length
online courses.

\begin{table}[H]
\begin{tabular}{|l|l|l|}
\hline
\textbf{Type}                      & \textbf{Number} & \textbf{\%} \\ \hline
Python (Jupyter or Colab Notebook) & 15 & 53.6        \\ \hline
Command line                       & 2  & 7.1         \\ \hline
Web page material                  & 11 & 39.3        \\ \hline
Video Lectures                     & 7  & 25          \\ \hline
Online Course / MOOC               & 4  & 14.3        \\ \hline
Powerpoint/Slides                  & 2  & 7.1         \\ \hline
\end{tabular}
\caption{Form of material}
\label{tab:format}
\end{table}

Though some of the resources are provided within the context of a course
or a workshop all the materials can be used self-directed and
self-paced. The suggested time commitment ranges from about 1 to up to
50 hours. Many of the materials are relatively short, most are a `one
shot' learning experience, i.e. half-day workshop; there are few
materials with a longer time horizon. The majority of the materials we
identified (n = 21) were published more recently (2019 - 2021). Only 7
resources were published between 2016 and 2018. This might underscore
the recent increase of interest in training accessible to a broader
audience.

\subsection{Content}

There is a potentially overwhelming amount of material focused on
teaching machine learning\footnote{As one example of this, a search for
  \hspace{0pt}\hspace{0pt}'machine+learning+introduction' in the
  `description' field of GitHub repositories returns 3024 results.
  Expanding this to
  \hspace{0pt}\hspace{0pt}'machine+learning+introduction' in the
  `readme' of GitHub repositories returns 88177 (search carried out
  2021/09/29)
  \href{https://gist.github.com/davanstrien/df2be4b0d5c5944f3f5974f3f1c4085e}{\url{https://gist.github.com/davanstrien/df2be4b0d5c5944f3f5974f3f1c4085e}}}.
Out of the material we reviewed, the scope ranges from complete MOOC
courses aiming to teach machine learning ``from scratch'' to blog posts
covering the details of one particular technique in great detail. Some
well known MOOCs include; Andrew Ng's Coursera Machine
Learning course \footnote{\url{https://www.coursera.org/learn/machine-learning}}, which has over 4 million people enrolled and the
fast.ai \href{https://course.fast.ai/}{{Practical Deep Learning
for Coders}} course and
\href{https://www.elementsofai.com/}{Elements of AI}, which has
the goal of educating 1\% of European Citizens on the basics of AI and
is notable for not focusing primarily on using coding examples or
exercises as a method for teaching machine learning. There is also a
growing trend for universities to share material for machine learning
courses openly shortly after or alongside the delivery of
courses.\footnote{For example Stanford:
  \href{http://web.stanford.edu/class/cs224n/}{http://web.stanford.edu/class/cs224n}
  and MIT:
  \href{http://introtodeeplearning.com/}{http://introtodeeplearning.com/}}
In addition, many machine learning frameworks and libraries offer
courses for working with their tools. For example, the spaCy NLP library
has an intro to \href{https://course.spacy.io/en/}{NLP course},
the
\href{https://scikit-learn.org/stable/index.html}{scikit-learn}
library has extensive
\href{https://scikit-learn.org/stable/tutorial/basic/tutorial.html}{tutorials}
and the \href{https://keras.io/}{Keras} library has a growing
number of code \href{https://keras.io/examples/}{examples}.

The materials in our review touched on a broad range of topics. Based on
the topics extracted from the materials the content broadly covers the
following topical areas:

\begin{itemize}
    \item Classification
    \item Clustering 
    \item Natural Language Processing 
    \item Word Vectors
    \item Digital Assistants 
    \item What is AI?
    \item Use cases
    \item Neural Networks, Deep Learning
    \item Ethics / societal implications of AI
    \item Text
    \item Images
    \item Machine learning terminology and concepts
    \item Data
    \item AI Project management
    \item Python
    \item APIs and Data collection
\end{itemize}

(for details see the
\href{https://docs.google.com/document/d/1pGRw5cU-NhFEQgrQgyAE35a0rvwU5PVOVi8maGeh670/edit?usp=sharing}{Training
Resources Content Map})

Not surprisingly there is a certain overlap between the materials.
Broadly speaking, they fall into the following categories:

\subsubsection{General Introductions to Machine Learning}

These workshops/materials aim to provide an introduction to machine
learning methods and potentially their particular application to Digital
Humanities (DH) and/or GLAMs. Examples include the
\href{https://carpentries.org/}{{Carpentries}} lesson
\href{https://github.com/carpentries-incubator/machine-learning-librarians-archivists}{Intro
to AI for GLAM} (under development), the webinar
\href{https://www.youtube.com/watch?v=vCjEVX0HGu8}{Artificial
Intelligence in the Library} and the
\href{https://github.com/sul-cidr/Workshops/tree/master/Intro_to_ML_with_Python}{Intro
to Machine Learning with Python} workshop materials. The high
percentage of materials that require Python or command line skills
indicate that the majority of the materials either focus on or require
coding. Notably,
\href{https://www.elementsofai.com/}{elementsofai.com} provides
an introduction to AI and ML concepts that does not require any coding.

\subsubsection{Introductions to a Specific Method or Tool}

These materials focus on introducing a particular method or specific
tools to do tasks, such as tutorials on Named Entity Recognition using
\href{https://github.com/impresso/named-entity-tutorial-dh2019/tree/master/notebooks}{spaCy}
or the
\href{https://github.com/YaleDHLab/lab-workshops/tree/master/named-entity-recognition}{CoreNLP
library}.

There are also tutorials supporting the use of particular machine
learning tools for GLAM sectors. For example,
\href{https://annif.org/}{Annif}, a tool for automatic subject
indexing, has associated
\href{https://github.com/NatLibFi/Annif-tutorial}{tutorials}
that demonstrate the use of the tool.\footnote{These types of tutorials may overlap with a simple documentation of the functionality.}

\subsubsection{Working with GLAM data}

A few of the courses focus on accessing GLAM data for use with ML.
Typically the focus is on a particular collection not on ML per se.
Examples are the \href{https://glam-workbench.net/}{GLAM
workbench}, for GLAM collections mainly from Australia and New Zealand,
the
\href{https://github.com/LibraryOfCongress/data-exploration}{data
explorer from the US Library of Congress}, and the
\href{https://github.com/CamLib/genizah-medical-visualisation}{Cambridge
Digital Library metadata}.

\subsection{Ethics}

We identified four sources that teach and address ethics in the context
of AI: \href{https://ethics-of-ai.mooc.fi/}{Ethics of AI},
\href{https://ethics.fast.ai/}{Practical Data Ethics} and
\href{https://course.fast.ai/videos/?lesson=5}{Lesson 5} of the
course \href{https://course.fast.ai/}{FastAI Deep Learning for
Coders}.
\href{https://fullstackdeeplearning.com/spring2021/lecture-9/}{Lesson
9} of \href{https://fullstackdeeplearning.com/}{Full Stack Deep
Learning}. These materials cover a comprehensive and wide range of
issues, including discrimination and bias, privacy and surveillance, accountability and fairness, explainability, as well as global and economic issues of AI, human rights and algorithmic colonialism.

\subsection{Audience}

The table below provides a summary of the training materials with GLAM
staff in mind as the target audience, use GLAM collections as training
data, or address applications and issues that are of particular
relevance for GLAM workflows and operations. Assuming some disciplinary
overlap between DH and GLAM, we included materials that had a focus on
DH and as such are potentially useful for a GLAM audience. DH materials
focused on topic modelling, for example, may focus on the use of topic
models to answer a research question, rather than the use of topic
models as a tool for information retrieval or collection management.
This suggests that relying solely on DH material for a curriculum may
lead to gaps in topic and method coverage.

\begin{table}[H]
\centering
\begin{tabular}{@{}ll@{}}
\toprule
\textbf{Focus}   & \textbf{Number} \\ \midrule
GLAM             & 12         \\
DH               & 5          \\
General audience & 11         \\ \bottomrule
\end{tabular}
\caption{Target audience}
\label{tab:Audience}
\end{table}

In our survey we were able to identify 17 resources (60.7\%) that have
relevance for GLAM, including DH focused material. Given that we
specifically set out to identify GLAM relevant materials this large
proportion may not be unexpected. The remaining 11 sources are
introductions for a general audience, though they tend to lean towards
an audience with some programming background.

Our search was aimed at materials for beginners i.e. they were
introductions to a topic rather than assuming some previous knowledge of
a topic and diving more deeply into one aspect of that topic. However,
whilst they were introductory in nature, they often assumed at least
some minimal Python knowledge and some familiarity with
\href{https://jupyter.org/}{Jupyter notebooks} for interacting
with code, thus limiting the audience to GLAM staff with a certain
familiarity of programming.

Although not aimed specifically at the GLAM sector, many of the not
explicitly DH or GLAM focused courses have the potential to be used for
teaching in the GLAM sector; the
\href{https://www.elementsofai.com/}{Elements of AI} and fastai
\href{https://course.fast.ai/}{course}, though not originally
designed with GLAM in mind, have both been used by GLAM staff study
groups.\footnote{\href{https://github.com/cncoleman/elementsofai4glam}{https://github.com/cncoleman/elementsofai4glam}
  \href{https://github.com/AI4LAM/fastai4GLAMS}{https://github.com/AI4LAM/fastai4GLAMS}}

\subsubsection{Pedagogy}

Broadly speaking, materials can be divided into those that are shared
as-is (``here it is, now you're on your own'') and courses, as for
example \href{https://www.deeplearning.ai/ai-for-everyone/}{AI
for Everyone}, specifically designed with pedagogical and learning
goals in mind.
\href{https://www.elementsofai.com/}{elementsofai} and
\href{https://buildingai.elementsofai.com/}{buildingai} have
support structures in place and are linked to a discussion forum where
participants engage with each other and the creators of the lessons.
\href{https://www.elementsofai.com/}{elementsofai} additionally
includes peer reviews of assigned short essays.

It is the nature of the web-based materials that they are generally
self-directed and to be used independently, which can be convenient for
some users. It is also the nature of web-based materials that they are
often not very interactive. Jupyter notebooks can remedy this, though
the interaction often does not go beyond executing and perhaps
experimenting with the code.

Two resources explicitly outline their teaching philosophies:
\href{https://carpentries.github.io/instructor-training/}{Carpentries}
and
\href{https://www.fast.ai/2016/10/08/teaching-philosophy/}{fast.ai}.

\subsubsection{Example Training Data}

Example ML training data used in the materials covered several types of
data. The table below provides an overview of those training resources
that included data\footnote{multiple types possible per resource}, which
are 22 within our collection. Over three quarters of the training data
are based on text, and about one third uses images. Tabular data and
audio visual materials are less represented.

\begin{table}[H]
\centering
\begin{tabular}{|l|l|l|}
\hline
\textbf{Type of data} & \textbf{Number}  & \textbf{\%}    \\ \hline
text         & 17 & 77.2 \\ \hline
image        & 8  & 36.4 \\ \hline
tabular      & 2  & 9.1  \\ \hline
audio, video & 1  & 4.5  \\ \hline
\end{tabular}
\caption{Data Formats used}
\label{tab:my-table}
\end{table}

Some of the reviewed materials use GLAM data, but many use ``generic''
machine learning datasets or synthetically generated `toy' data.

A few of the lessons use GLAM data, for example
\href{https://programminghistorian.org/en/lessons/OCR-and-Machine-Translation}{OCR
and Machine Translation} uses data from the Wilson Center Digital
Archive's collection on Iran-Soviet relations.

There is also a growing body of materials that aim to provide
introductions to working with particular GLAM collections, like
\href{https://glam-workbench.github.io/}{https://glam-workbench.github.io/},
\href{https://github.com/BL-Labs/Jupyter-notebooks-projects-using-BL-Sources}{https://github.com/BL-Labs/Jupyter-notebooks-projects-using-BL-Sources},
and
\href{https://github.com/LibraryOfCongress/data-exploration}{https://github.com/LibraryOfCongress/data-exploration}.
Whilst these materials are not always focused on machine learning
directly they may help lower barriers to accessing collections and cover
some of the required preprocessing steps for working with GLAM data for
machine learning tasks.

The training data have limitations for teaching and learning in the GLAM
sector:

\begin{itemize}
\item It is often tidier than would be found in the `real world'; for
  example there are no missing labels, or the labels are equally
  distributed between classes.
\item It may have labels that are not as relevant for GLAM applications,
  making it harder for learners to translate the material to their own
  work.
\item The results may be better than would be achieved on real-world data,
  leading to unreasonable expectations of performance.
\end{itemize}

\subsubsection{Maintenance}

Based on information from the website or GitHub commits we tried to
determine when the materials were last updated. An overview is provided
in the table below.

\begin{table}[H]
\centering
\begin{tabular}{!{\color{black}\vrule}l!{\color{black}\vrule}l!{\color{black}\vrule}l!{\color{black}\vrule}} 
\hline
\textbf{Year published} & \textbf{Year of last update} & \textbf{Number }  \\ 
\hline
2016                    & 2016                         & 1                 \\ 
\hline
2017                    & 2020                         & 1                 \\ 
\hline
2017                    & 2021                         & 1                 \\ 
\hline
2018                    & 2018                         & 2                 \\ 
\hline
2018                    & 2019                         & 1                 \\ 
\hline
2018                    & 2020                         & 1                 \\ 
\hline
2019                    & 2019                         & 5                 \\ 
\hline
2019                    & 2020                         & 1                 \\ 
\hline
2019                    & 2021                         & 2                 \\ 
\hline
2020                    & 2020                         & 5                 \\ 
\hline
2020                    & 2021                         & 3                 \\ 
\hline
2021                    & 2021                         & 5                 \\
\hline
\end{tabular}
\caption{Year published and year last updated}
\label{tab:Year of publication vs last updated}
\end{table}

With regard to evaluating maintenance, we face the issue of later
materials being less in need of updating. About half of the earlier
materials were updated at least until the following year, while the
other half of the materials published 2019 and earlier (8 out of the 15)
were not kept updated in later years. We observed some link rot with
datasets for some lessons.

Creators of the materials are affiliated with academic institutions,
libraries, private companies (fastai), or are software developers
(scikit learn). It appears that often material was prepared for a
particular workshop or event and not updated subsequently. This is not
necessarily a problem, but as software libraries change and machine
learning develops, it may be desirable to have material updated
semi-regularly over time. In a similar vein, we considered whether the
material was 'community developed', i.e. developed and maintained by a
larger group with a straightforward process for contributing\footnote{This
  was easier to assess for materials hosted on GitHub where
  contributions are more visible. For material hosted on Github the
  average number of contributors was 6, however this is skewed by a few
  materials having large number of contributors (maximum was 37). The
  mode value was 1.}. Whilst community developed materials are by no
means always desirable, there may be duplication of efforts by creating
some similar material independently. There may also be more scope for
continued improvement, review and refinement of materials when developed
collectively.

\section{Recommendations for teaching and learning AI for
GLAM}

The greatest value for teaching and learning AI will come from a GLAM
specific focus on the unique issues faced by GLAM institutions,
targeting GLAM use cases and institutional contexts. Below are the
priorities we have identified, many of which do not appear in the
materials reviewed.

\subsection{There is a need for machine learning concepts
(without
coding)}

As has been discussed above, there are many important parts of the
Machine Learning pipeline that do not require or involve writing code.
This includes many important steps with implications for the success of
machine learning projects, but also a conceptual understanding of the
process and how the models are trained. We recommend that the AI4LAM
community seek to ensure that not all materials focus on writing code as
the primary way of engaging with machine learning. Whilst some of this
material might be introductory in nature it is also useful to have
non-introductory material focused on more `advanced' or specialised
topics which don't include coding.

To consider training and skill building for the various GLAM roles more
specifically we developed a number of
\href{https://docs.google.com/document/d/1Qvf31mQklZvBdVt_iRaIIE7VQU1FMjz3Kb4UORm9Lug/edit?usp=sharing}{`stick
figure personas'}\footnote{A persona is a sketch, usually based on user
  research, of current or potential user types for a particular
  application/system/etc. These are intended to help make the assessment
  of systems or in this case teaching materials easier in relation to
  the needs of these different personas.}. The audiences we think are
not addressed adequately are shaped by the personas discussed above. For
example, one of the personas is `Metadata M' who
\href{https://docs.google.com/document/d/1Qvf31mQklZvBdVt_iRaIIE7VQU1FMjz3Kb4UORm9Lug/edit\#}{``works
as a metadata librarian at a large academic library and has joined a
campus-wide AI initiative. M is enthusiastic to learn as much as
possible about AI so they can be an active contributor to the group.
They don't expect to do much coding, but would like to try.''} Keeping
this type of persona in mind we can identify the lack of materials that
don't focus on coding as a problem.

Whilst many parts of the machine learning pipeline and machine learning
projects involve writing code, this is not the only part of such
projects and potentially not even the most critical part of the process.
For example, clearly defining the business use case for machine
learning, including how machine learning will integrate into workflows,
might be more important than a specific method or tool being used. These
skills often require domain expertise and some broad understanding of
machine learning and its limitations but often don't require code to be
written.

\subsection{Where coding is necessary, python is important}

Python and Jupyter Notebooks are standard tools for machine learning and
thus constitute a large part of the training material, even if
introductory. Even in GLAM focused materials, some Python knowledge is
often required. This raises the question if introductory Python and
Jupyter Notebooks training should be provided for GLAM staff before
teaching basic ML concepts and if so, how that could or should be done,
particularly for those without coding background. On the other hand, we
should consider and explore alternative approaches for hands-on
approaches to teaching AI, that are interactive and intuitive and do not
rely on coding.

\subsection{Community building and project-based work will improve learning.}

Much of the AI training material is oriented to teaching machine
learning to engineers and developers. As a result, learners are
typically left to their own devices to work through it and be
resourceful in solving problems or answering questions they might have.
Often the material is presented as step-by-step instructions or Jupyter
notebooks to follow along.

As we have noted, integrating AI into the GLAM sector has implications
for many different roles and areas of expertise beyond engineering. It
will involve changes to our practices and processes. If we are to move
from learning machine learning to learning how to integrate machine
learning methods into our practices and processes, a collaborative
learning experience is necessary, with GLAM engineers and coders working
side-by-side with subject experts, curators, and others.

\subsection{Maintenance and discovery of training materials for the GLAM sector will benefit from community organizing.}

We identified a certain extent of duplication of effort and the
potential risks of material not being updated. This raises the issue of
sustained support structures for training materials. There may be some
scope for developing a shared approach to creating, reviewing and
maintaining teaching materials via community development, review and
maintenance strategies. This is a significant undertaking and would
require broad community support.

There are examples to draw from, for example the
\href{https://carpentries.org/}{Carpentries} and
\href{https://programminghistorian.org/}{Programming Historian},
both of which have maintained a growing body of teaching materials for
significant time periods. A fist step would be to further explore the
benefits and feasibility of developing these shared resources and
assessing potential models for undertaking this type of project.

When researching materials for this report we followed a variety of
pointers. There is no one stop shop for this kind of material. The
\href{https://github.com/CENL-Network-Group-AI/awesome-list)}{CENL
Awesome List} is one example we found of curating a directory of GLAM
focused machine learning resources. While offering a potential route to
increasing discoverability, long-term maintenance of this type of list
may be challenging.

Alternatively, there are potentially `quick wins' to aid discover that
don't require as much maintenance:

\begin{itemize}
\item
  \begin{quote}
  Applying \href{https://github.com/topics/glam}{GLAM} or
  other similar \href{https://github.com/topics}{topics} to
  materials shared on GitHub would make it easier to disambiguate
  GLAM-focused material from other material, and for example help
  distinguish a search for `machine learning libraries' (which provides
  an introduction to Named Entity Recognition for historic text) from
  `machine learning libraries' (which provide an introduction to
  Pytorch).
  \end{quote}
\item
  \begin{quote}
  Considering Search Engine Optimization when sharing teaching
  materials. There is a large volume of blogs, notebooks, and YouTube
  videos related to machine learning and data science. Relying solely on
  `generic' machine learning keywords may limit how many people discover
  your material.
  \end{quote}
\end{itemize}

\subsection{An emphasis on critical data practices and the
ethical implications of Machine Learning is necessary}

Data is a central component of machine learning but isn't always given
the same attention as models. Since GLAM institutions already play such
a central role in the collection, description, organization,
transformation and dissemination of data, it is potentially even more
important that data is given a prominent part in a GLAM-focused AI
curriculum. This becomes even more important when considering the
potential ethical ramifications of using data held by GLAM institutions.
Strategies, tools and workflows for building training data, also include
questions of provenance, metadata, and labeling and how to produce
datasheets \footnote{\url{https://arxiv.org/abs/1803.09010}} for GLAM
data used for training ML models.

Much of the material we reviewed focused on building the `model' part of
the machine learning pipeline. Often materials start with a prepared
training set (often with a pre-created training/validation/test split).
The data collection and preparation stages of the ML pipeline, such as
sampling, choosing labels, and annotating data are too often left out.
There is little attention given to the processes involved to actually
get there: how to generate training data, decide labels, inter-annotator
agreement etc., and the issues around these, that involve awareness of
the biases and ethical implications of that process.

In addition, while there are several materials addressing ethics
generally, none of them address ethics in the GLAM context. Given the
nature of GLAM collections and the particular awareness around ethical
issues already existing in GLAM organizations, it is of particular
importance to consider the implications for machine learning and steps
which can be taken to mitigate ethical risks.

The process of creating machine learning models involves a range of
different stages including (but not limited to):

\begin{itemize}
\item Collecting data
\item Sampling and processing data
\item Creating training sets including:
\begin{itemize}
  \item Choosing (appropriate) labels
  \item Identifying existing metadata schema to integrate with 
  \item Creating training, validation and test splits
  \item Identifying biases in data
  \end{itemize}
\item Tracking the provenance of various models and datasets
\item Integrating models into tools or systems
\item Integrating or making use of predictions as part of existing or new
  workflows
\item Monitoring data drift and model performance overtime
\item Documenting models and data
\item Communication to external and internal audiences
\end{itemize}


The broader data science community is also focusing increasingly on the
data component of machine learning\footnote{For example
  \href{https://https-deeplearning-ai.github.io/data-centric-comp/}{https://https-deeplearning-ai.github.io/data-centric-comp/}}
so there are likely to be potential synergies with non GLAM focused
initiatives that will help with the development of data focused teaching
materials.

\subsection{GLAM specific training data will improve teaching and learning.}

Data is a central component of machine learning but isn't always given the same attention as models. Since GLAM institutions already play such a central role in the collection, description, organization, transformation and dissemination of data, it is potentially even more important that data is given a prominent part in a GLAM-focused AI curriculum.

The lack of training data has been identified as a barrier to adopting
machine learning in libraries\footnote{\href{https://labs.loc.gov/static/labs/work/reports/Cordell-LOC-ML-report.pdf}{https://labs.loc.gov/static/labs/work/reports/Cordell-LOC-ML-report.pdf}}.
This is not only a problem of implementation, it is also a problem for
teaching and learning. Using GLAM specific data is crucial for helping
the learner see the relevance of AI to their work. It offers
opportunities to discuss challenges GLAM projects might face, e.g. poor
OCR, grayscale images, historical language etc.

Time and money are barriers to developing training data both in academia
and industry. Training materials focused on this topic may help lower
the barrier. The process of developing these datasets requires subject
expertise and skills in data manipulation.

\subsection{Effective implementation requires a focus on teaching
the design and management of ML projects}

None of the materials identified in this study address issues around the
implementation and management of ML projects in GLAM organizations.
Issues to be addressed should include:

\begin{itemize}
\item Communicating the results of machine learning to other staff and the
  public
\item How to collect annotated data for machine learning
 \item How to document training data (for example writing datasheets \footnote{\href{https://arxiv.org/abs/1803.09010}{datasheets}{https://arxiv.org/abs/1803.09010}}
\item How to document models (for example writing model cards\footnote{\href{https://arxiv.org/pdf/1810.03993.pdf}{https://arxiv.org/pdf/1810.03993.pdf}}
 \item How to evaluate model performance (in particular, in relation to a
  specific GLAM task or use-case as opposed to `out of the box'
  evaluation metrics)
 \item \href{https://en.wikipedia.org/wiki/Software_deployment}{Deployment}'
  of ML in libraries and how this might not always match industry
  approaches
\item Communicating with end-users about the use of machine learning, e.g.
  how to share metadata produced through machine learning methods with
  end-users and how much disclaimer needs to be attached etc.
\item Examples for the entire workflow of a ML project, that would move from
  collection through an ML process and interpretation
\end{itemize}

GLAM projects that share materials associated with both successful, and
importantly, unsuccessful projects would contribute to learning
resources. Topics that such material would cover are:

\begin{itemize}
\item Why was the project chosen? How does it fit into a broader
  institutional context?
\item What approach was taken by the project? Why was machine learning
  chosen as opposed to another approach?
\item What data was available? How it was gathered? How were labels chosen?
  Who did the annotation? Were there challenges in shaping the
  annotation guidelines?, etc.
\item How was the model `deployed' or used by the institution?
\item What challenges, barriers and failures were encountered along the way?
\end{itemize}

Another model, suggested by Cordell (2020)\footnote{Machine Learning +
  Libraries} is that of ``Implementation Toolkits'':

\begin{quote}
``The specific workflows outlined in the existing literature can be hard
to generalize from, particularly for teams new to ML work. To that end,
funders and leading ML libraries should support the development of ML
Implementation Toolkits for distinct data types (e.g. text, image,
audio, video) or in particular domains. These toolkits would include:

1. Model training data,with descriptions of annotation processes
undertaken and an intellectual justification of the same.

2. An inventory of existing resources, including open-access,
domain-specific collections; ML models, algorithms, and code that could
be used or adapted; and prospective pre-training data from earlier ML
projects in the domain.

3. A walkthrough of the full technical pipeline from related project,
including training, valida- tion, and application of the model.

4. Model code from related experiments.'' (Cordell, 2020, p 58)
\end{quote}

These toolkits would cover a broader range of the pipeline and would
serve as a guide for institutions wishing to work with machine learning.

We would challenge, however, the idea of these toolkits being produced
only by "leading" libraries or GLAM institutions. Whilst some
institutions may invest more in machine learning and should be
encouraged to share lessons and guidance based on this experience, we
believe broad sharing of lessons learned across the GLAM sector would be
beneficial. For example, the lessons learned in one setting, a national
library, may not be as relevant to a local archive service.

\section{Conclusion}

While this review of materials, methods and target audiences for AI
training is a snapshot from the period of time in which it was compiled,
it brings to light a significant trend toward both upskilling and
reskilling within the GLAM community. Though not all GLAM resources are
digitized and many may never be digitized, the training reviewed here
can be applicable to GLAM work at the level of metadata. At the same
time, as the amount of digitized and born digital cultural heritage
continues to grow, it seems abundantly clear that GLAM practices will
need to adopt and adapt techniques from data science and, specifically,
machine learning in order to keep pace. To this end, this review offers
guidance for the GLAM community about current offerings and future
directions.


\end{document}